%% file: main.tex
\documentclass[10pt,twocolumn,letterpaper]{article}

\usepackage[pagenumbers]{cvpr} 

\input{preamble}
\definecolor{cvprblue}{rgb}{0.21,0.49,0.74}
\definecolor{lncolor}{HTML}{FF8DA8}

\usepackage{colortbl}
\usepackage{array}
\usepackage{amsmath}
\usepackage{multirow}
\usepackage{makecell}
\usepackage{multicol}
\usepackage{graphicx}
\usepackage{algorithmicx,algorithm}
\usepackage[dvipsnames,svgnames]{xcolor}
\usepackage{color}
\usepackage{xcolor}
\usepackage{svg}
\usepackage{alltt}
\usepackage{caption}
\usepackage{subcaption}

\usepackage{pifont}
\definecolor{goodgreen}{rgb}{0.0, 0.5, 0.0}
\definecolor{badred}{rgb}{0.7, 0.0, 0.0}

\usepackage[pagebackref,breaklinks,colorlinks,allcolors=cvprblue]{hyperref}

\def\paperID{4900} 
\def\confName{CVPR}
\def\confYear{2026}

\title{From Points to Clouds: Learning Robust Semantic Distributions for Multi-modal Prompts}

\author{Weiran Li \hspace{0.2cm}
	Yeqiang Liu \hspace{0.2cm}
    Yijie Wei \hspace{0.2cm}
	Mina Han \hspace{0.2cm}
	Xin Liu \hspace{0.2cm}
	Zhenbo Li \thanks{Corresponding author.}
	\vspace{1.5mm}\\
	China Agricultural University
	\vspace{1mm}\\
	{\tt\small vranlee86@gmail.com, lizb@cau.edu.cn}\\
    {\tt\small yeqiangliu@cau.edu.cn, yjwei@cau.edu.cn, hmnsx666@163.com, lyc@cau.edu.cn}
}

\begin{document}
\maketitle
\input{sec/0_abstract}    
\input{sec/1_intro}
\input{sec/2_related}
\input{sec/3_method}
\input{sec/4_exps}
\input{sec/5_conclusion}

\input{sec/X_suppl}

{
	\small
	\bibliographystyle{ieeenat_fullname}
	\bibliography{main}
}

\end{document}

%% file: sec/0_abstract.tex
\begin{abstract}
Multimodal Prompt Learning (MPL) has emerged as a pivotal technique for adapting large-scale Visual Language Models (VLMs). However, current MPL methods are fundamentally limited by their optimization of a single, static point representation. This paradigm is inherently brittle, leads to overfitting on base classes, and generalizes poorly to novel or ambiguous categories. We challenge this point paradigm, proposing that robust generalization requires learning a semantic cloud (i.e., a distribution over the embedding space).
To achieve this, we introduce \textbf{Points-to-Clouds (P2C)}, a novel framework inspired by diffusion models that reframes prompt learning as a dynamic denoising task. At the core of P2C is a dual denoising mechanism: a Dynamic Prompt Denoising (DPD) mechanism perturbs text prompts with sophisticated, annealed noise to learn a smoother semantic landscape, while an auxiliary V-L Mapper denoising loss re-tasks the mapper as a denoising autoencoder. This forces the mapper to reconstruct clean visual prompts from noisy text inputs, ensuring robust cross-modal alignment. Extensive experiments across 11 datasets demonstrate that P2C consistently outperforms strong baselines. On the base-to-novel generalization benchmark, our method attains a harmonic mean of 79.7\%, representing a relative improvement of 1.4\% over the baseline. The code and models are available in the \textit{\href{https://vranlee.github.io/P2C/}{https://vranlee.github.io/P2C/}}.
\end{abstract}

%% file: sec/1_intro.tex
\section{Introduction}
\label{sec:intro}

\begin{figure}[t]
    \centering
    \includegraphics[width=\linewidth]{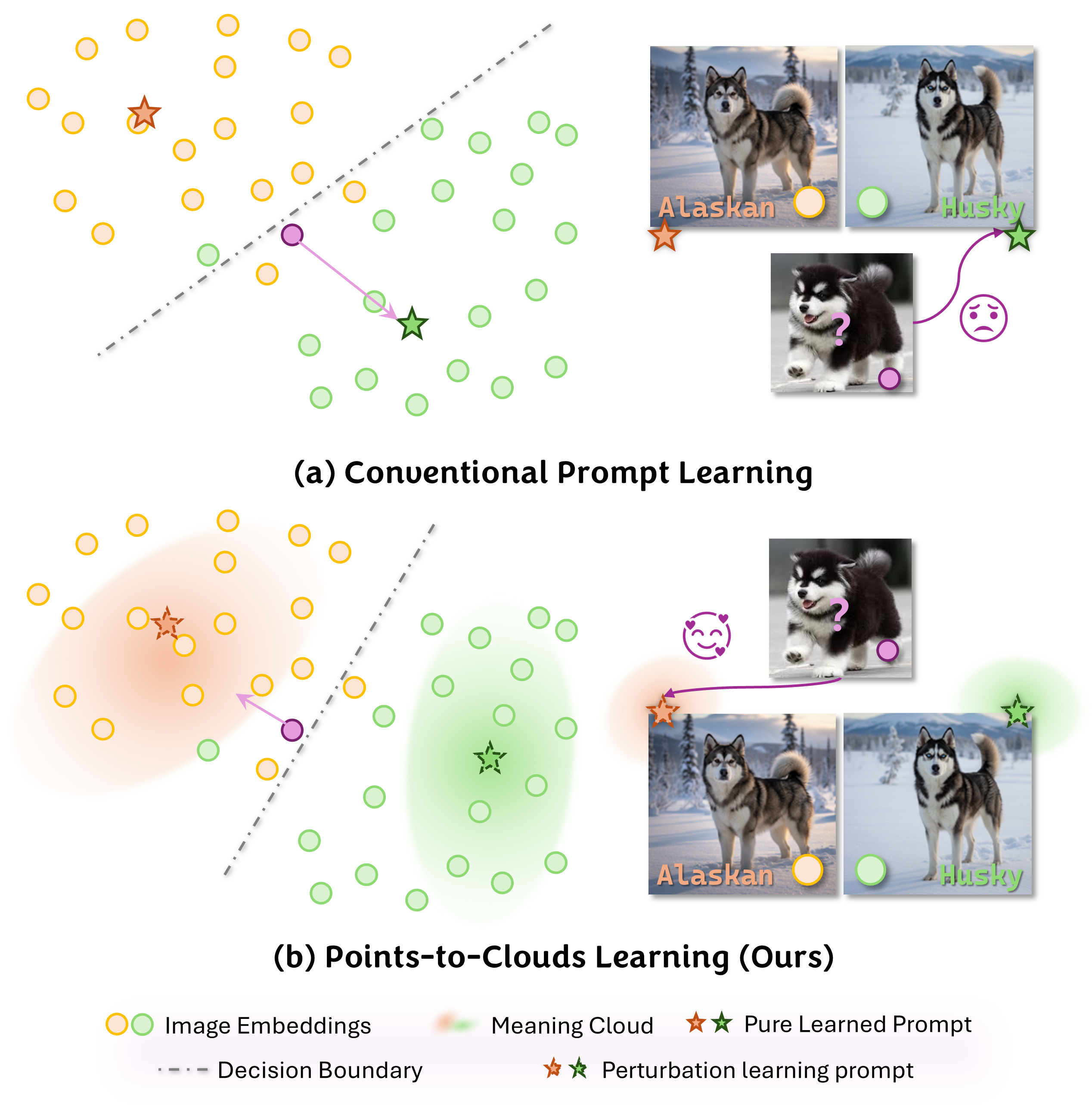}
    \caption{\textbf{(a)} Conventional prompt learning optimizes for a static \textit{point} representation. This approach is brittle and overfits, creating fragile decision boundaries (e.g., misclassifying an \textit{Alaskan} puppy as a \textit{Husky}). \textbf{(b)} Our \textit{Points-to-Clouds} (P2C) learns a robust \textit{semantic cloud}. It captures a smoother semantic region, enabling the correct classification of ambiguous, out-of-distribution samples.}
    \label{fig:concept}
\end{figure}

Large-scale Vision-Language Models (VLMs)~\cite{radford2021learning}, such as CLIP~\cite{radford2021learning}, have shown remarkable zero-shot generalization capabilities by learning aligned representations from vast amounts of web-scale image-text data. Parameter-Efficient Prompt Learning (PEPL)~\cite{zhou2022learning} has emerged as a dominant paradigm for adapting these frozen VLMs to downstream tasks. By learning only a few task-specific vectors (prompts) while keeping the massive model weights frozen, these methods offer an efficient alternative to full fine-tuning.

Despite its success, the pioneering prompt learning method~\cite{zhou2022learning} quickly revealed a core challenge: its learned static prompts severely overfit the training (base) classes, leading to a dramatic performance drop on unseen (novel) categories~\cite{zhou2022conditional}. To mitigate this base-to-novel trade-off, subsequent research bifurcated into two main streams. One stream focuses on retaining pre-trained knowledge through consistency regularization, compelling the learnable prompt's output to align with that of a hand-crafted prompt~\cite{yao2023visual, zhu2023prompt, khattak2023self}. A parallel stream introduced conditional mechanisms to make prompts dynamic, such as generating instance-specific prompts~\cite{zhou2022conditional}.

More recently, state-of-the-art methods have explored more sophisticated architectures. For instance, MaPLe~\cite{khattak2023maple} proposed multi-modal deep prompting, injecting learnable prompts deep into both vision and language encoders and linking them via a projection function. Concurrently, ATPrompt~\cite{li2024atprompt} identified the category-centric nature of prompts as a limitation, proposing an attribute-anchored design that uses universal attributes as a semantic bridge to enhance alignment with unknown categories.

Despite their sophistication, these advanced methods still share a fundamental limitation inherited from their predecessors: they all aim to optimize a single, static representation vector for each prompt, i.e., a \textit{point} in the high-dimensional embedding space. As illustrated in Fig.~\ref{fig:concept}(a), this point-based paradigm is inherently brittle. Such optimization is highly sensitive to the training data distribution, making it prone to overfitting on spurious correlations within the few-shot training data. This results in a fragile decision boundary that fails when encountering ambiguous or out-of-distribution samples (e.g., an \textit{Alaskan} puppy being misclassified as a \textit{Husky}).

In this paper, we challenge this point-based paradigm. We argue that true robustness requires learning a semantic distribution, or a \textit{cloud} (Fig.~\ref{fig:concept}(b)), rather than an isolated \textit{point}. We posit that this \textit{cloud} captures a robust semantic region resilient to input variations. Inspired by the remarkable success of diffusion models~\cite{ho2020denoising} in learning complex data distributions by reversing a noising process, we reframe multi-modal prompt learning as a dynamic denoising task. Instead of optimizing a static point, we train our model to find a robust semantic cloud by challenging it to denoise perturbed prompts. Our approach, \textit{Points-to-Clouds} (P2C), operationalizes this idea through a dual denoising mechanism. Specifically, we employ a \textit{Dynamic Prompt Denoising} (DPD) mechanism, which perturbs the learnable text prompts (both class and attribute prompts) during training.
This process is meticulously controlled, going beyond simple noise injection~\cite{pan2025nlprompt} by using sophisticated noise modeling combined with an annealing schedule. This schedule progressively reduces the noise intensity, guiding the model from coarse-grained to fine-grained learning. Moreover, we introduce an auxiliary V-L Mapper denoising loss. This loss explicitly re-tasks the V-L Mapper (the projection function between modalities) as a denoising autoencoder. It is trained to reconstruct clean visual prompt representations from noisy text prompt inputs. This auxiliary task forces the V-L Mapper to learn a deep and robust semantic alignment, understanding the underlying semantic correspondence rather than a superficial, surface-level mapping. This dual mechanism effectively trains the model to navigate a smooth and robust semantic space, thereby learning the semantic cloud we propose.

Our contributions are summarized as follows:
\begin{itemize}
    \item We identify the static, point-based representation as a fundamental limitation of existing prompt learning methods, one that leads to brittle overfitting and poor generalization. We challenge this paradigm and propose the \textit{Points-to-Clouds} (P2C) framework that learns a robust semantic distribution (or \textit{cloud}) to enhance generalization and out-of-distribution robustness.

    \item We devise a novel, diffusion-inspired dual denoising mechanism to learn this semantic cloud. This features: (i) a \textit{Dynamic Prompt Denoising} (DPD) strategy that uses sophisticated, annealed noise to guide the model in learning a smoother semantic landscape; and (ii) an auxiliary V-L Mapper denoising loss that re-tasks the V-L Mapper as a denoising autoencoder, forcing it to learn deep semantic alignments rather than shallow mappings.

    \item Extensive experiments across 11 datasets demonstrate that P2C achieves a 79.7\% Harmonic Mean on the challenging base-to-novel generalization benchmark and demonstrates strong robustness in cross-dataset and domain generalization tasks.
\end{itemize}

%% file: sec/2_related.tex
\section{Related Work}
\label{sec:related}

\paragraph{Prompt Learning in VLMs.}
Prompt learning originated in NLP and was first adapted to VLMs by CoOp~\cite{zhou2022learning}.
CoOp replaces hand-crafted text prompts (e.g., \textit{a photo of a \{class\}}) with a set of learnable continuous vectors, demonstrating strong performance on few-shot tasks.
However, CoOp's static prompts were shown to overfit severely to the (\textit{base}) classes seen during training and failed to generalize to unseen (\textit{novel}) classes~\cite{zhou2022conditional}.
To address this limitation, CoCoOp~\cite{zhou2022conditional} introduced an instance-conditional approach, using a lightweight network to generate a dynamic prompt vector for each input image, which significantly improved novel class generalization.

\paragraph{Improving Generalization.}
Following CoOp and CoCoOp, a significant line of research has focused on mitigating the trade-off between base-class specialization and novel-class generalization.
One popular strategy is to preserve the pre-trained knowledge of the frozen VLM by introducing a regularization or consistency loss.
Methods like KgCoOp~\cite{yao2023visual}, ProGrad~\cite{zhu2023prompt}, and PromptSRC~\cite{khattak2023self} all enforce, in different ways, that the learnable prompt's output remains consistent with the output of the original hand-crafted prompt.
A parallel approach treats this as a knowledge distillation problem, where a student prompt model learns from a frozen teacher, either from the VLM itself~\cite{li2024promptkd} or a larger, senior model~\cite{wu2024cascade}.
While effective, these methods can introduce a strong bias towards the pre-trained knowledge, a bias that DeKg~\cite{li2025divergence} attempts to balance by adding a divergence-promoting loss.

\paragraph{Multi-modal and Structured Prompts.}
Another line of work identified the text-only approach of CoOp and CoCoOp as a bottleneck.
MaPLe~\cite{khattak2023maple} pioneered multi-modal prompt learning by introducing learnable prompts into the deep layers of both the vision and language encoders.
Crucially, MaPLe links them via a V-L Mapper, forcing the two modalities to adapt synergistically.
This multi-modal design has become a foundation for many state-of-the-art methods~\cite{yang2024mma, zheng2025hierarchical}.
Concurrently, researchers have explored more structured prompt semantics.
ATPrompt~\cite{li2024atprompt} and FATE~\cite{xu2025fate} proposed using explicit, fine-grained attributes as anchors within the prompt.
Other works, like TCP~\cite{yao2024tcp} and BIP~\cite{yao2025bi}, generate dynamic, class-aware prompts by mapping frozen class embeddings.
Others have explored decoupling prompts into components for base \textit{vs.} novel tasks~\cite{li2025dpc} or foreground \textit{vs.} background~\cite{zhang2025decouple}.

\paragraph{From Points to Clouds.}
Despite these advances in architectural and semantic design, all existing methods still optimize for a single, deterministic point representation for their prompts.
We argue this is the core reason for their inherent brittleness.
In this work, we propose a novel paradigm: learning a robust semantic distribution, or a \textit{cloud}, instead of a single \textit{point}.
While the simple injection of noise to improve robustness has been explored~\cite{pan2025nlprompt}, our work formalizes this idea through a diffusion-inspired framework.
We introduce a dual denoising mechanism that not only perturbs text prompts with sophisticated, annealed noise but also explicitly re-tasks the MaPLe-style V-L Mapper as a denoising autoencoder to learn a robust, distributed semantic mapping.

%% file: sec/3_method.tex
\section{Methodology}
\label{sec:method}

This work introduces a diffusion-inspired training framework for learning robust, distributed prompt representations.
We build upon a strong baseline that synergizes multi-modal deep prompting with attribute anchoring.
We first review this baseline and then detail our proposed dual denoising mechanism.

\subsection{Preliminary}

Our method builds upon a baseline that synergizes two state-of-the-art methods: MaPLe~\cite{khattak2023maple} and ATPrompt~\cite{li2024atprompt}.
\begin{itemize}
    \item \textbf{MaPLe:} We adopt the MaPLe architecture, which learns deep prompts in both the vision and text encoders. A crucial component is the V-L Mapper, denoted as $F$, which projects text-side prompts to create corresponding visual-side prompts. This ensures that adaptations in one modality are coherently mirrored in the other.
    \item \textbf{ATPrompt:} We adopt the ATPrompt design for the text encoder. Instead of learning only class-specific tokens, the text prompt is structured as an attribute-category hybrid. It consists of a set of learnable vectors $P_{\text{cls}}$ (for the class) and $P_{\text{att}}$ (for the attributes), anchored by fixed universal attribute tokens (e.g., [color], [shape]). 
\end{itemize}

The baseline text prompt, $P_{\text{base}}$, thus combines these elements. While strong, this baseline still optimizes for a single, static set of prompt vectors $\{P_{\text{cls}}, P_{\text{att}}\}$, which we identify as a \textit{point} representation.

\subsection{Points-to-Clouds (P2C)}

\begin{figure*}[htbp]
\centering
    \includegraphics[width=1\linewidth]{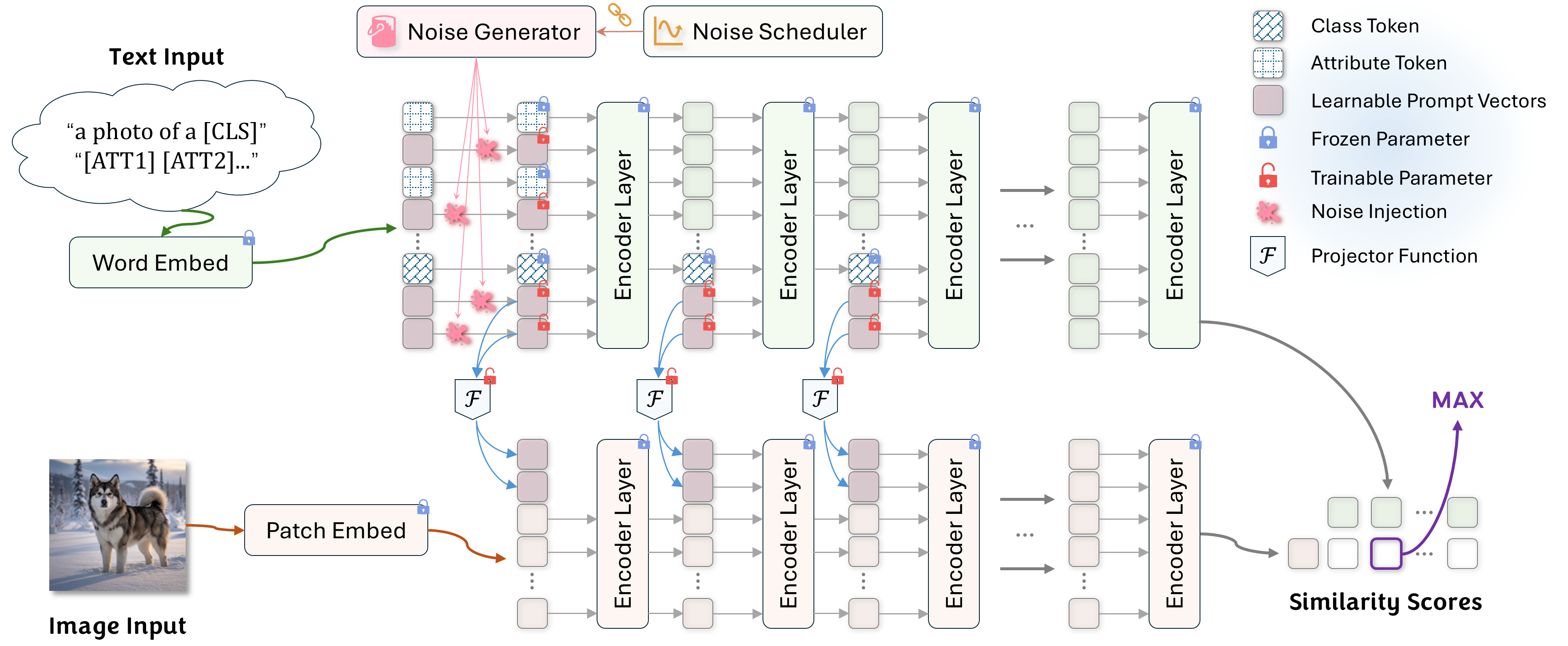}
    \caption{\textit{Points-to-Clouds} (P2C) framework. The dual denoising mechanism learns a robust semantic \textit{cloud} instead of a \textit{point}: text prompts are perturbed with annealed noise, while the V-L Mapper (\textit{F}) is trained to reconstruct clean visual prompts from the noisy inputs.}
 \label{fig:pipeline}
\end{figure*}

As established in Section \ref{sec:intro}, optimizing for a static \textit{point} representation $P$ is inherently brittle and prone to overfitting (Figure~\ref{fig:concept}(a)).
We hypothesize that a truly robust model must learn a semantic \textit{cloud}, or a distribution $p(P)$, which captures a resilient semantic region (Fig.~\ref{fig:concept}(b)).

Inspired by diffusion models~\cite{ho2020denoising}, we propose to learn this distribution by reframing the optimization objective.
Instead of minimizing the loss for a single prompt $P$, we optimize the expectation of the loss over a distribution of noisy prompts.
The optimization task is thus transformed from learning a static vector to a dynamic denoising task.
This is achieved through our dual denoising mechanism, which targets both the prompt representations and the V-L Mapper.

\subsection{Dynamic Prompt Denoising (DPD)}
\label{sec:method_denoising}

This mechanism targets the primary classification objective.
During each training step, we perturb the learnable prompt vectors before they are fed into the text encoder.
Let $P_t = \{P_{\text{cls}}, P_{\text{att}}\}$ be the set of learnable text prompts. We generate a noisy version $\tilde{P}_t$ as:
\begin{equation}
\tilde{P}_t = P_t + \epsilon_t
\end{equation}
where $\epsilon_t$ is a structurally-matched noise vector sampled at the current training step $t$.
The model then performs classification using this perturbed prompt $\tilde{P}_t$.
This process, which we term \textit{Dynamic Prompt Denoising} (DPD), forces the model to learn a smoother semantic landscape and prevents it from relying on brittle, overfitted features of a single prompt vector.

The effectiveness of this process is governed by two key components:

\subsubsection{Sophisticated Noise Modeling}
Instead of applying simple standard Gaussian noise ($\epsilon \sim \mathcal{N}(0, I)$), we posit that the target semantic cloud is likely multi-modal, not unimodal.
We therefore employ a Gaussian Mixture Model (GMM) as our noise generator.
This allows the noise $\epsilon_t$ to be sampled from a more complex, multi-modal distribution, challenging the model with more structured perturbations that better simulate complex variations within the semantic embedding space.

\subsubsection{Annealed Denoising Schedule}
The magnitude of the injected noise is critical. We introduce a noise scheduler that anneals the noise intensity over the course of training, guided by the current epoch $t$ and total epochs $T$.
The noise scale $\sigma_t$ is calculated by a schedule function $S(t, T)$:
\begin{equation}
\sigma_t = \sigma_{\text{max}} \cdot S(t, T)
\end{equation}
This strategy, directly inspired by diffusion models, begins with a high noise level ($\sigma_t \approx \sigma_{\text{max}}$), forcing the model to learn coarse-grained, robust features first. As training progresses, $\sigma_t$ decreases, allowing the model to fine-tune the prompts on more subtle details.

\subsection{Auxiliary V-L Mapper Denoising}
\label{sec:method_aux_loss}

A central component of our P2C framework is to explicitly train the V-L Mapper $F$ as a denoising autoencoder. This constitutes a new, auxiliary training task.

The original purpose of $F$ in the MaPLe baseline is to map clean text prompts $P_t$ to clean visual prompts $P_v$.
We retain this objective but force $F$ to learn it under duress. We define an auxiliary consistency loss $L_{\text{aux}}$ as follows:
\begin{align}
P_v^{\text{clean}} &= F(P_t) \label{eq:clean_vis} \\
\tilde{P}_t &= P_t + \eta_t \label{eq:noisy_text} \\
P_v^{\text{recon}} &= F(\tilde{P}_t) \label{eq:recon_vis} \\
L_{\text{aux}} &= \text{MSE}(P_v^{\text{recon}}, P_v^{\text{clean}}) \label{eq:aux_loss}
\end{align}
where $\eta_t$ is a noise vector sampled independently for this auxiliary task.

This loss explicitly trains the V-L Mapper $F$ to reconstruct the \textit{clean} visual prompt representation (Eq.~\ref{eq:clean_vis}) from a \textit{noisy} text prompt input (Eq.~\ref{eq:noisy_text}).
This forces $F$ to learn a deep, robust semantic correspondence between the two modalities, rather than a brittle, superficial mapping.
It must understand the underlying semantic cloud to successfully denoise the input.

\subsection{Overall Training Objective}
Our final training objective $L_{\text{total}}$ combines the standard classification loss $L_{\text{cls}}$ with our new auxiliary denoising loss $L_{\text{aux}}$.

The main classification loss is computed using the dynamically noised prompts from our DPD mechanism (Section \ref{sec:method_denoising}):
\begin{equation}
L_{\text{cls}} = \text{CrossEntropy}(f(I, \tilde{P}_t), y)
\end{equation}
The total loss $L_{\text{total}}$ is a weighted sum of these two objectives:
\begin{equation}
L_{\text{total}} = L_{\text{cls}} + \lambda \cdot L_{\text{aux}}
\end{equation}
where $\lambda$ is a hyperparameter that balances the two tasks.
This dual-objective framework trains the model to be both an accurate classifier and a robust denoiser, effectively guiding the optimization process to find the robust semantic cloud.

%% file: sec/4_exps.tex
\section{Experiments}
\label{sec:exps}

\begin{table*}[ht!]
\caption{Experiments in \emph{base-to-novel} generalization with the ViT-B/16 visual backbone. All methods use $k{=}16$ shots per base class.}
\label{tab:b2n}
    \def\arraystretch{1.075}
    \centering
    \scriptsize
    \begin{minipage}{0.3\textwidth}
        \centering
        \caption*{\textit{Average}}\vspace{-1em}
        \begin{tabular}{lcc|c}
        \toprule
        \textbf{Method} & \textbf{Base} & \textbf{Novel} & \textbf{HM} \\
        \midrule
        CLIP \cite{radford2021learning} & 69.3 & 74.2 & 71.7 \\
        CoOp \cite{zhou2022learning} & 82.7 & 63.2 & 71.7 \\
        CoCoOp \cite{zhou2022conditional} & 80.5 & 71.7 & 75.8 \\
        ProGrad \cite{zhu2023prompt} & 82.5 & 70.8 & 76.2 \\
        KgCoOp \cite{yao2023visual} & 80.7 & 73.6 & 77.0 \\
        MaPLe \cite{khattak2023maple} & 82.3 & 75.1 & 78.6 \\
        \midrule
        \rowcolor{lncolor!30}
        P2C (Ours) & \textbf{83.5} & \textbf{76.1} & \textbf{79.7} \\
        \bottomrule
        \end{tabular}
        \vspace{0.5em}
    \end{minipage}%
    \hfill
    \centering
    \begin{minipage}{0.3\textwidth}
        \centering
        \caption*{ImageNet}\vspace{-1em}
        \begin{tabular}{lcc|c}
        \toprule
        \textbf{Method} & \textbf{Base} & \textbf{Novel} & \textbf{HM} \\
        \midrule
        CLIP \cite{radford2021learning} & 72.4 & 68.1 & 70.2 \\
        CoOp \cite{zhou2022learning} & 76.5 & 67.9 & 71.9 \\
        CoCoOp \cite{zhou2022conditional} & 76.0 & 70.4 & 73.1 \\
        ProGrad \cite{zhu2023prompt} & 77.0 & 66.7 & 71.5 \\
        KgCoOp \cite{yao2023visual} & 75.8 & 70.0 & 72.8 \\
        MaPLe \cite{khattak2023maple} & 76.7 & 70.5 & 73.5 \\
        \midrule
        \rowcolor{lncolor!30}
        P2C (Ours) & \textbf{77.1} & \textbf{70.9} & \textbf{73.9} \\
        \bottomrule
        \end{tabular}
        \vspace{0.5em}
    \end{minipage}%
    \hfill
    \centering
    \begin{minipage}{0.3\textwidth}
        \centering
        \caption*{Caltech101}\vspace{-1em}
        \begin{tabular}{lcc|c}
        \toprule
        \textbf{Method} & \textbf{Base} & \textbf{Novel} & \textbf{HM} \\
        \midrule
        CLIP \cite{radford2021learning} & 96.8 & 94.0 & 95.4 \\
        CoOp \cite{zhou2022learning} & 98.0 & 89.8 & 93.7 \\
        CoCoOp \cite{zhou2022conditional} & 98.0 & 93.8 & 95.8 \\
        ProGrad \cite{zhu2023prompt} & 98.0 & 93.9 & 95.9 \\
        KgCoOp \cite{yao2023visual} & 97.7 & 94.4 & 96.0 \\
        MaPLe \cite{khattak2023maple} & 97.7 & 94.4 & 96.0 \\
        \midrule
        \rowcolor{lncolor!30}
        P2C (Ours) & \textbf{98.3} & \textbf{95.1} & \textbf{96.7} \\
        \bottomrule
        \end{tabular}
        \vspace{0.5em}
    \end{minipage}%
    
    \vfill

    \begin{minipage}{0.3\textwidth}
        \centering
        \caption*{Flowers102}\vspace{-1em}
        \begin{tabular}{lcc|c}
        \toprule
        \textbf{Method} & \textbf{Base} & \textbf{Novel} & \textbf{HM} \\
        \midrule
        CLIP \cite{radford2021learning} & 72.1 & \textbf{77.8} & 74.8 \\
        CoOp \cite{zhou2022learning} & 97.6 & 59.7 & 74.1 \\
        CoCoOp \cite{zhou2022conditional} & 94.9 & 71.8 & 81.7 \\
        ProGrad \cite{zhu2023prompt} & 95.5 & 71.9 & 82.0 \\
        KgCoOp \cite{yao2023visual} & 95.0 & 74.7 & 83.7 \\ 
        MaPLe \cite{khattak2023maple} & 95.9 & 72.5 & 82.6 \\
        \midrule
        \rowcolor{lncolor!30}
        P2C (Ours) & \textbf{97.9} & 75.2 & \textbf{85.1} \\
        \bottomrule
        \end{tabular}
        \vspace{0.5em}
    \end{minipage}
    \hfill
    \begin{minipage}{0.3\textwidth}
        \centering
        \caption*{Oxford Pets}\vspace{-1em}
        \begin{tabular}{lcc|c}
        \toprule
        \textbf{Method} & \textbf{Base} & \textbf{Novel} & \textbf{HM} \\
        \midrule
        CLIP \cite{radford2021learning} & 91.2 & 97.3 & 94.1 \\
        CoOp \cite{zhou2022learning} &  93.7 & 95.3 & 94.5 \\
        CoCoOp \cite{zhou2022conditional} & 95.2 & 97.7 & 96.4 \\
        ProGrad \cite{zhu2023prompt} & 95.1 & 97.6 & 96.3 \\
        KgCoOp \cite{yao2023visual} & 94.7 & \textbf{97.8} & 96.2 \\
        MaPLe \cite{khattak2023maple} & 95.4 & \textbf{97.8} & 96.6 \\
        \midrule
        \rowcolor{lncolor!30}
        P2C (Ours) & \textbf{96.2} & 97.7 & \textbf{96.9} \\
        \bottomrule
        \end{tabular}
        \vspace{0.5em}
    \end{minipage}%
    \hfill
    \begin{minipage}{0.3\textwidth}
        \centering
        \caption*{Stanford Cars}\vspace{-1em}
        \begin{tabular}{lcc|c}
        \toprule
        \textbf{Method} & \textbf{Base} & \textbf{Novel} & \textbf{HM} \\
        \midrule
        CLIP \cite{radford2021learning} & 63.4 & 74.9 & 68.7 \\
        CoOp \cite{zhou2022learning} & \textbf{78.1} & 60.4 & 68.1 \\
        CoCoOp \cite{zhou2022conditional} & 70.5 & 73.6 & 72.0 \\
        ProGrad \cite{zhu2023prompt} & 77.7 & 68.6 & 72.9 \\
        KgCoOp \cite{yao2023visual} & 71.8 & \textbf{75.0} & 73.4 \\
        MaPLe \cite{khattak2023maple} & 72.9 & 74.0 & 73.5 \\
        \midrule
        \rowcolor{lncolor!30}
        P2C (Ours) & 76.9 & 73.3 & \textbf{75.1} \\
        \bottomrule
        \end{tabular}
        \vspace{0.5em}
    \end{minipage}%
    
    \vfill

    \begin{minipage}{0.3\textwidth}
        \centering
        \caption*{Food101}\vspace{-1em}
        \begin{tabular}{lcc|c}
        \toprule
        \textbf{Method} & \textbf{Base} & \textbf{Novel} & \textbf{HM} \\
        \midrule
        CLIP \cite{radford2021learning} & 90.1 & 91.2 & 90.7 \\
        CoOp \cite{zhou2022learning} & 88.3 & 82.3 & 85.2 \\
        CoCoOp \cite{zhou2022conditional} & \textbf{90.7} & 91.3 & 91.0 \\
        ProGrad \cite{zhu2023prompt} & 90.4 & 89.6 & 90.0 \\
        KgCoOp \cite{yao2023visual} & 90.5 & 91.7 & 91.1 \\
        MaPLe \cite{khattak2023maple} & \textbf{90.7} & \textbf{92.1} & \textbf{91.4} \\
        \midrule
        \rowcolor{lncolor!30}
        P2C (Ours) & \textbf{90.7} & 92.0 & 91.3 \\
        \bottomrule
        \end{tabular}
        \vspace{0.5em}
    \end{minipage}
    \hfill
    \begin{minipage}{0.3\textwidth}
        \centering
        \caption*{FGVC Aircraft}\vspace{-1em}
        \begin{tabular}{lcc|c}
        \toprule
        \textbf{Method} & \textbf{Base} & \textbf{Novel} & \textbf{HM} \\
        \midrule
        CLIP \cite{radford2021learning} & 27.2 & \textbf{36.3} & 31.1 \\
        CoOp \cite{zhou2022learning} & 40.4 & 22.3 & 28.8 \\
        CoCoOp \cite{zhou2022conditional} & 33.4 & 23.7 & 27.7 \\
        ProGrad \cite{zhu2023prompt} & \textbf{40.5} & 27.6 & 32.8 \\
        KgCoOp \cite{yao2023visual} & 36.2 & 33.6 & 34.8 \\
        MaPLe \cite{khattak2023maple} & 37.4 & 35.6 & 36.5 \\
        \midrule
        \rowcolor{lncolor!30}
        P2C (Ours) & 39.0 & 34.9 & \textbf{36.8} \\
        \bottomrule
        \end{tabular}
        \vspace{0.5em}
    \end{minipage}%
    \hfill
    \begin{minipage}{0.3\textwidth}
        \centering
        \caption*{SUN397}\vspace{-1em}
        \begin{tabular}{lcc|c}
        \toprule
        \textbf{Method} & \textbf{Base} & \textbf{Novel} & \textbf{HM} \\
        \midrule
        CLIP \cite{radford2021learning} & 69.4 & 75.4 & 72.2 \\
        CoOp \cite{zhou2022learning} & 80.6 & 65.9 & 72.5 \\
        CoCoOp \cite{zhou2022conditional} & 79.7 & 76.9 & 78.3 \\
        ProGrad \cite{zhu2023prompt} & 81.3 & 74.2 & 77.6 \\
        KgCoOp \cite{yao2023visual} & 80.3 & 76.5 & 78.4 \\
        MaPLe \cite{khattak2023maple} & 80.8 & \textbf{78.7} & 79.8 \\
        \midrule
        \rowcolor{lncolor!30}
        P2C (Ours) & \textbf{81.7} & 78.4 & \textbf{80.0} \\
        \bottomrule
        \end{tabular}
        \vspace{0.5em}
    \end{minipage}
    
    \vfill 

    \begin{minipage}{0.3\textwidth}
        \centering
        \caption*{DTD}\vspace{-1em}
        \begin{tabular}{lcc|c}
        \toprule
        \textbf{Method} & \textbf{Base} & \textbf{Novel} & \textbf{HM} \\
        \midrule
        CLIP \cite{radford2021learning} & 53.2 & \textbf{59.9} & 56.4 \\
        CoOp \cite{zhou2022learning} & 79.4 & 41.2 & 54.2 \\
        CoCoOp \cite{zhou2022conditional} & 77.0 & 56.0 & 64.9 \\
        ProGrad \cite{zhu2023prompt} & 77.4 & 52.4 & 62.5 \\
        KgCoOp \cite{yao2023visual} & 77.6 & 55.0 & 64.4 \\
        MaPLe \cite{khattak2023maple} & \textbf{80.4} & 59.2 & \textbf{68.2} \\
        \midrule
        \rowcolor{lncolor!30}
        P2C (Ours) & 80.3 & 58.1 & 67.4 \\
        \bottomrule
        \end{tabular}
        \vspace{0.5em}
    \end{minipage}%
    \hfill
    \begin{minipage}{0.3\textwidth}
        \centering
        \caption*{EuroSAT}\vspace{-1em}
        \begin{tabular}{lcc|c}
        \toprule
        \textbf{Method} & \textbf{Base} & \textbf{Novel} & \textbf{HM} \\
        \midrule
        CLIP \cite{radford2021learning} & 56.5 & 64.1 & 60.0 \\
        CoOp \cite{zhou2022learning} & 92.2 & 54.7 & 68.7 \\
        CoCoOp \cite{zhou2022conditional} & 87.5 & 60.0 & 71.2 \\
        ProGrad \cite{zhu2023prompt} & 90.1 & 60.9 & 72.7 \\
        KgCoOp \cite{yao2023visual} & 85.6 & 64.3 & 73.5 \\
        MaPLe \cite{khattak2023maple} & 94.1 & 73.2 & 82.4 \\
        \midrule
        \rowcolor{lncolor!30}
        P2C (Ours) & \textbf{96.6} & \textbf{84.3} & \textbf{90.0} \\
        \bottomrule
        \end{tabular}
        \vspace{0.5em}
    \end{minipage}
    \hfill
    \begin{minipage}{0.3\textwidth}
        \centering
        \caption*{UCF101}\vspace{-1em}
        \begin{tabular}{lcc|c}
        \toprule
        \textbf{Method} & \textbf{Base} & \textbf{Novel} & \textbf{HM} \\
        \midrule
        CLIP \cite{radford2021learning} & 70.5 & 77.5 & 73.9 \\
        CoOp \cite{zhou2022learning} & \textbf{84.7} & 56.1 & 67.5 \\
        CoCoOp \cite{zhou2022conditional} & 82.3 & 73.5 & 77.6 \\
        ProGrad \cite{zhu2023prompt} & 84.3 & 74.9 & 79.4 \\
        KgCoOp \cite{yao2023visual} & 82.9 & 76.7 & 79.7 \\
        MaPLe \cite{khattak2023maple} & 83.0 & \textbf{78.7} & \textbf{80.8} \\
        \midrule
        \rowcolor{lncolor!30}
        P2C (Ours) & 84.2 & 77.6 & \textbf{80.8} \\
        \bottomrule
        \end{tabular}
        \vspace{0.5em}
    \end{minipage}
\end{table*}

\subsection{Datasets}
Our methodology is rigorously evaluated across a comprehensive suite of 11 diverse datasets, adhering to the standard benchmarks established in prior works~\cite{zhou2022learning, khattak2023maple}. This suite encompasses a wide range of visual recognition tasks: generic object recognition (\textit{e.g.,} ImageNet~\cite{deng2009imagenet} and Caltech101~\cite{fei2004learning}); fine-grained visual classification (Oxford Pets~\cite{parkhi2012cats}, Stanford Cars~\cite{krause20133d}, Flowers102~\cite{nilsback2008automated}, Food101~\cite{bossard2014food}, and FGVC Aircraft~\cite{maji2013fine}); as well as specialized tasks such as scene recognition (SUN397~\cite{xiao2010sun}), texture classification (DTD~\cite{cimpoi2014describing}), satellite imagery analysis (EuroSAT~\cite{helber2019eurosat}), and action recognition (UCF101~\cite{soomro2012ucf101}). For the domain generalization task, we adopt the standard protocol of training on ImageNet (the source domain) and evaluating on four out-of-distribution variants as target domains: ImageNetV2~\cite{recht2019imagenet}, ImageNet-Sketch~\cite{wang2019learning}, ImageNet-A~\cite{hendrycks2021natural}, and ImageNet-R~\cite{hendrycks2021many}.

Our evaluation is conducted under three standard protocols: 
\textbf{(1) Base-to-Novel Generalization.} For each dataset, we partition the classes into \textit{base} (seen) and \textit{novel} (unseen) sets. Models are trained in a 16-shot setting on the base classes and are evaluated on the test sets of both base and novel classes. Performance is benchmarked using the accuracy on base classes (Base), novel classes (Novel), and their Harmonic Mean (HM).
\textbf{(2) Cross-Dataset Transfer.} Models are first trained on ImageNet with 16 shots per class. They are then evaluated on the other 10 datasets in a zero-shot transfer setting.
\textbf{(3) Domain Generalization.} Following the cross-dataset setup, the 16-shot ImageNet-trained model is directly evaluated on the four ImageNet variants to assess its robustness against domain shifts.

\subsection{Implementation Details}
All experiments are conducted using the ViT-B/16 CLIP backbone. We optimize the models using Stochastic Gradient Descent (SGD) with a batch size of 4. The learning rate is set to $3.5 \times 10^{-3}$ for the base-to-novel generalization benchmark, and $2.6 \times 10^{-3}$ for the cross-dataset and domain generalization tasks. To ensure robust and reproducible results, each experiment is repeated three times with different random seeds, and we report the average performance.

\subsection{Comparison with State-of-the-Art Methods} 
\label{sec:exp_sota}

\subsubsection{Base-to-Novel Generalization}

A comprehensive evaluation of base-to-novel generalization is presented in Table~\ref{tab:b2n}. Our P2C framework achieves a strong Harmonic Mean (HM) of 79.7\%, consistently outperforming the baseline methods. This performance indicates that our dual denoising mechanism effectively enhances generalization ability while preserving model specialization on base classes.

The benefits of our approach are particularly pronounced on challenging fine-grained and specialized datasets. For example, on EuroSAT, P2C yields a significant 7.6\% improvement in HM over the MaPLe baseline. On the fine-grained datasets Flowers102 and Stanford Cars, it also achieves notable gains of 2.5\% and 1.6\%, respectively. Furthermore, on general benchmarks like ImageNet and Caltech101, P2C demonstrates a consistent advantage over MaPLe and shows highly competitive performance on diverse datasets such as Food101 and UCF101.

These results strongly support our central hypothesis: shifting the learning paradigm from a static point representation to a dynamic semantic cloud is a key strategy for mitigating the critical trade-off between overfitting and generalization in Vision-Language Models.

\subsubsection{Domain Generalization}

\begin{table}[t]
\centering
\caption{Comparison of P2C with previous state-of-the-art methods on domain generalization across 4 datasets.}
\label{tab:domain_generalization}
\footnotesize
\begin{tabular}{rc|cccc|c} 
\toprule
 & Source & \multicolumn{5}{c}{Target} \\ \cmidrule(l){2-7} 
 & ImageNet & -V2 & -S & -A & -R & Avg. \\ \cmidrule(l){2-7}
CLIP \cite{radford2021learning} & 66.7 & 60.8 & 46.2 & 47.8 & 74.0 & 57.2 \\ 
CoOp \cite{zhou2022learning} & \textbf{71.5} & 64.2 & 48.0 & 49.7 & 75.2 & 59.3 \\
CoCoOp \cite{zhou2022conditional} & 71.0 & 64.1 & 48.8 & 50.6 & 76.2 & 59.9 \\ 
PromptSRC \cite{khattak2023self} & 71.3 & 64.4 & \textbf{49.6} & \textbf{50.9} & \textbf{77.8} & \textbf{60.7} \\
MMA \cite{yang2024mma} & 71.0 & 64.3 & 49.1 & 51.1 & 77.3 & 60.5 \\ 
MaPLe \cite{khattak2023maple} & 70.7 & 64.1 & 49.2 & \textbf{50.9} & 77.0 & 60.3 \\ 
\midrule
\rowcolor{lncolor!30} P2C (Ours) & 71.2 & \textbf{64.6} & 49.1 & 50.4 & 77.5 & 60.4 \\ 
\bottomrule
\end{tabular}
\vspace{-0.4cm}
\end{table}

To further probe the robustness of our framework, we evaluated P2C on demanding domain generalization tasks. As detailed in Table~\ref{tab:domain_generalization}, a single model trained on ImageNet was deployed in a zero-shot manner to four distinct out-of-distribution target datasets.

P2C achieves a competitive average accuracy of 60.4\% across the four target domains. This performance is on par with recent methods like PromptSRC and MMA, and notably outperforms the MaPLe baseline, all without compromising fidelity on the source domain (71.2\% on ImageNet). We attribute this robustness to our framework's core design. Unlike point-based paradigms that learn fragile representations overfitted to source-specific cues (\textit{e.g.}, artistic style or texture), P2C's diffusion-inspired denoising guides the model to learn a distributed semantic cloud. This cloud captures the invariant, core semantics of a category, resulting in representations that are inherently more stable against distributional shifts.

\subsubsection{Cross-Dataset Transfer}

\begin{table*}[t]
\centering
\caption{Comparison of P2C with previous state-of-the-art methods on cross-dataset evaluation across 10 datasets.}
\label{tab:cross_dataset}
    \footnotesize
    \begin{tabular}{rc|ccccccccccc}
    \toprule
     &
     Source &
     \multicolumn{11}{c}{Target} \\ \cmidrule(l){2-13} 
     &
     \rotatebox{60}{ImageNet} &
     \rotatebox{60}{\textit{Average}} &
     \rotatebox{60}{Caltech101} &
     \rotatebox{60}{Oxford Pets} &
     \rotatebox{60}{Stanford Cars} &
     \rotatebox{60}{Flowers102} &
     \rotatebox{60}{Food101} &
     \rotatebox{60}{FGVC Aircraft} &
     \rotatebox{60}{SUN397} &
     \rotatebox{60}{DTD} &
     \rotatebox{60}{EuroSAT} &
     \rotatebox{60}{UCF101} \\ \cmidrule(l){2-13} 
    CoOp \cite{zhou2022learning} &
     \textbf{71.5} &
     63.9 &
     93.7 &
     89.1 &
     64.5 &
     68.7 &
     85.3 &
     18.5 &
     64.2 &
     41.9 &
     46.4 &
     66.6 \\
    CoCoOp \cite{zhou2022conditional} &
     71.0 &
     65.7 &
     \textbf{94.4} &
     90.1 &
     65.3 &
     71.9 &
     86.1 &
     22.9 &
     67.4 &
     45.7 &
     45.4 &
     68.2 \\
    PromptSRC \cite{khattak2023self} &
     71.3 &
     65.8 &
     93.6 &
     90.3 &
     65.7 &
     70.3 &
     86.2 &
     23.9 &
     67.1 &
     \textbf{46.9} &
     45.5 &
     \textbf{68.8} \\
    TCP \cite{yao2024tcp} &
     71.4 &
     66.3 &
     94.0 &
     \textbf{91.3} &
     64.7 &
     71.2 &
     \textbf{86.7} &
     23.5 &
     67.2 &
     44.4 &
     \textbf{51.5} &
     68.7 \\
    MMA \cite{yang2024mma} &
     71.0 &
     \textbf{66.6} &
     93.8 &
     90.3 &
     \textbf{66.1} &
     72.1 &
     86.1 &
     \textbf{25.3} &
     \textbf{68.2} &
     46.6 &
     49.2 &
     68.3 \\
    MaPLe \cite{khattak2023maple} &
     70.7 &
     66.3 &
     93.5 &
     90.5 &
     65.6 &
     \textbf{72.2} &
     86.2 &
     24.7 &
     67.0 &
     46.5 &
     48.1 &
     68.7 \\\midrule
     \rowcolor{lncolor!30} P2C (Ours) &
     71.2&
     65.9&
     93.8&
     89.7&
     65.2&
     69.5&
     86.2&
     24.9&
     67.3&
     44.4&
     49.5&
     68.6 \\ \bottomrule
    \end{tabular}
\vspace{-0.3cm}
\end{table*}

We evaluated P2C's zero-shot transfer capabilities by training it on ImageNet and testing it on 10 unseen datasets (Table~\ref{tab:cross_dataset}). The framework achieves an average accuracy of 65.9\%, placing its performance in a comparable range to several recent state-of-the-art methods. This result aligns with the central thesis of our work. The \textit{point-to-cloud} paradigm, operationalized by our dual denoising mechanism, is designed to learn a generalizable semantic landscape rather than overfitting to the specific features of the source domain. The benefit of this approach is particularly evident on datasets with significant domain shifts; for instance, P2C maintains robust performance on challenging targets like EuroSAT (49.5\%) and FGVC Aircraft (24.9\%), validating that our method indeed captures more transferable, core semantics.

\subsection{Ablation Studies}
\label{sec:exp_ablation}

\begin{table}[t]
    \centering
    \caption{Ablation study on the main components of our P2C framework. We incrementally add Dynamic Prompt Denoising (DPD) and the Auxiliary V-L Mapper Denoising loss ($L_{\text{aux}}$). GM and GMM denote the noise model used for DPD.}
    \label{tab:ablation_components}
        \footnotesize
        \begin{tabular}{lccccc}
            \toprule
            \textbf{Method} & \textbf{DPD} & \textbf{$L_{\text{aux}}$} & \textbf{HM} & \textbf{$\Delta$} & \textbf{$\Delta$\%} \\
            \midrule
            Baseline & \ding{55} & \ding{55} & 77.0 & - & - \\
            \midrule
            + DPD (GM) & \ding{51} & \ding{55} & 78.3 & +1.3 & +1.69\% \\
            + DPD (GMM) & \ding{51} & \ding{55} & 78.3 & +1.3 & +1.69\% \\
            \midrule
            + DPD (GM) + $L_{\text{aux}}$ & \ding{51} & \ding{51} & 77.4 & +0.4 & +0.52\% \\
            \midrule       
            \rowcolor{lncolor!30}
            P2C (Ours) & \ding{51} & \ding{51} & \textbf{78.5} & 
            \textbf{+1.5}& 
            \textbf{+1.95\%} \\
            
            \bottomrule
        \end{tabular}
\end{table}

We conduct a comprehensive ablation study to isolate the contribution of each component within the P2C framework, with results detailed in Table~\ref{tab:ablation_components}. Our analysis starts from a strong baseline that synergizes multi-modal deep prompting (MaPLe-style) with attribute anchoring (ATPrompt-style), establishing a performance of 77.0\% Harmonic Mean (HM).

First, integrating the Dynamic Prompt Denoising (DPD) mechanism alone yields a significant performance boost, elevating the HM to 78.3\% (+1.3\%). This provides direct evidence for our core hypothesis: reframing prompt learning as a denoising task is the primary driver for learning a more robust and generalizable semantic cloud.

Subsequently, incorporating the auxiliary V-L mapper denoising loss ($L_{\text{aux}}$) completes the full P2C framework and further lifts the HM to 78.5\% (+0.2\%). This incremental gain demonstrates that $L_{\text{aux}}$ acts as an effective regularizer. It promotes a more robust cross-modal alignment by compelling the V-L mapper to reconstruct clean visual representations from noisy textual inputs, thereby refining the quality of the learned semantic cloud.

\subsection{Radical}

\begin{table}[t]
    \centering
    \caption{Radical analysis of core hyperparameters. Each row isolates the effect of a single parameter on the model's performance, with all others held at their optimal values (highlighted).
      \textbf{Sched.}: Noise Annealing Scheduler,
      \textbf{$\sigma_{\text{max}}$}: Max. noise std. for DPD,
      \textbf{$\sigma_{\text{aux}}$}: Noise std. for V-L Mapper,
      \textbf{$\lambda$}: Loss weight for V-L Mapper,
      \textbf{$s_H$}: Hidden scale factor for V-L Mapper.}
    \label{tab:ablation_combined_wide}
        \footnotesize
        \begin{tabular}{lcccccc}
            \toprule
            \textbf{Varied Setting} & \textbf{Sched.} & \textbf{$\sigma_{\text{max}}$} & \textbf{$\sigma_{\text{aux}}$} & \textbf{$\lambda$} & \textbf{$s_H$} & {\textbf{HM}} \\
            \midrule
            \multirow{4}{*}{\textit{Scheduler}} 
            & Constant & 0.015 & 0.01 & 0.1 & 2 & 78.2 \\
            & Linear & 0.015 & 0.01 & 0.1 & 2 & 78.0 \\
            & Cosine & 0.015 & 0.01 & 0.1 & 2 & 78.0 \\
            & \cellcolor{lncolor!30}Sigmoid & \cellcolor{lncolor!30}0.015 & \cellcolor{lncolor!30}0.01 & \cellcolor{lncolor!30}0.1 & \cellcolor{lncolor!30}2 & \cellcolor{lncolor!30}\textbf{78.7} \\
            \midrule
            \multirow{5}{*}{\textit{$\sigma_{\text{max}}$}} 
            & Sigmoid & 0.005 & 0.01 & 0.1 & 2 & 77.3 \\
            & Sigmoid & 0.01 & 0.01 & 0.1 & 2 & 76.5 \\
            & \cellcolor{lncolor!30}Sigmoid & \cellcolor{lncolor!30}0.015 & \cellcolor{lncolor!30}0.01 & \cellcolor{lncolor!30}0.1 & \cellcolor{lncolor!30}2 & \cellcolor{lncolor!30}\textbf{78.7} \\
            & Sigmoid & 0.02 & 0.01 & 0.1 & 2 & 77.0 \\
            & Sigmoid & 0.03 & 0.01 & 0.1 & 2 & 75.4 \\
            \midrule
            \multirow{3}{*}{\textit{$\sigma_{\text{aux}}$}} 
            & \cellcolor{lncolor!30}Sigmoid & \cellcolor{lncolor!30}0.015 & \cellcolor{lncolor!30}0.01 & \cellcolor{lncolor!30}0.1 & \cellcolor{lncolor!30}2 & \cellcolor{lncolor!30}\textbf{78.7} \\
            & Sigmoid & 0.015 & 0.02 & 0.1 & 2 & 78.3 \\
            & Sigmoid & 0.015 & 0.1 & 0.1 & 2 & 77.8 \\
            \midrule
            \multirow{8}{*}{\textit{$\lambda$}} 
            & Sigmoid & 0.015 & 0.01 & -1.0 & 2 & 77.5 \\
            & Sigmoid & 0.015 & 0.01 & -0.5 & 2 & 78.4 \\
            & Sigmoid & 0.015 & 0.01 & -0.2 & 2 & 78.3 \\
            & Sigmoid & 0.015 & 0.01 & -0.1 & 2 & 78.0 \\
            & \cellcolor{lncolor!30}Sigmoid & \cellcolor{lncolor!30}0.015 & \cellcolor{lncolor!30}0.01 & \cellcolor{lncolor!30}0.1 & \cellcolor{lncolor!30}2 & \cellcolor{lncolor!30}\textbf{78.7} \\
            & Sigmoid & 0.015 & 0.01 & 0.2 & 2 & 77.7 \\
            & Sigmoid & 0.015 & 0.01 & 0.5 & 2 & 77.5 \\
            & Sigmoid & 0.015 & 0.01 & 1.0 & 2 & 77.2 \\
            \midrule
            \multirow{6}{*}{\textit{$s_H$}} 
            & Sigmoid & 0.015 & 0.01 & 0.1 & 1 & 77.4 \\
            & \cellcolor{lncolor!30}Sigmoid & \cellcolor{lncolor!30}0.015 & \cellcolor{lncolor!30}0.01 & \cellcolor{lncolor!30}0.1 & \cellcolor{lncolor!30}2 & \cellcolor{lncolor!30}\textbf{78.7} \\
            & Sigmoid & 0.015 & 0.01 & 0.1 & 4 & 78.6 \\
            & Sigmoid & 0.015 & 0.01 & 0.1 & 6 & 77.5 \\
            & Sigmoid & 0.015 & 0.01 & 0.1 & 8 & 77.6 \\
            & Sigmoid & 0.015 & 0.01 & 0.1 & 16 & 75.9 \\
            \bottomrule
        \end{tabular}
\end{table}

\begin{figure*}[htbp]
\centering
    \includegraphics[width=1\linewidth]{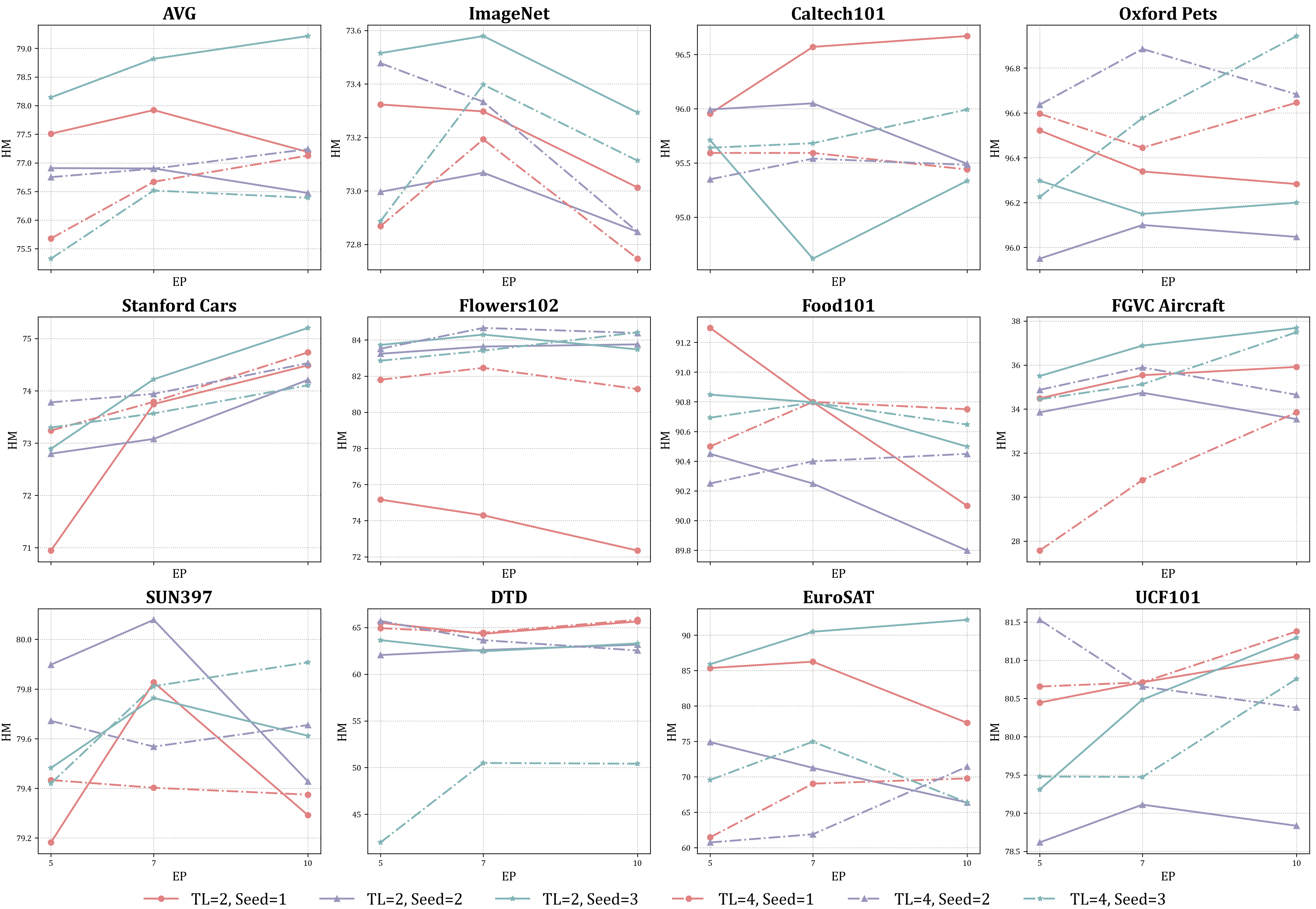}
    \caption{Performance under different attribute Token Length (TL) and different seeds.}
 \label{fig:pipeline}
\end{figure*}

To validate our design principles, we conduct a radical analysis of P2C's core hyperparameters in Table~\ref{tab:ablation_combined_wide}, dissecting three fundamental aspects that underpin its success.

\paragraph{Learning Dynamics: The Annealing Scheduler.}
The superiority of the diffusion-inspired Sigmoid scheduler (78.5\% HM) over simpler strategies confirms that a coarse-to-fine annealing curriculum is essential. This dynamic process—guiding the model from learning broad features under high noise to refining details under low noise—is critical for steering optimization away from fragile local optima.
\paragraph{Perturbation Intensity: The Noise Sweet Spot.}
Our analysis reveals a clear "sweet spot" for perturbation intensity at $\sigma_{\text{max}}=0.015$ (78.7\% HM). This highlights a critical balance, as insufficient noise fails to shift the model from point-based learning, while excessive noise disrupts convergence. This positions noise magnitude as the fundamental driving force for exploring the semantic cloud.
\paragraph{Balancing Objectives: The Auxiliary Task.}
Finally, the optimal parameters for the auxiliary task ($\lambda=0.1$, $s_H=2$) reveal its role as a carefully balanced regularizer. This balance requires precisely calibrating the loss weight $\lambda$ to provide meaningful regularization, while ensuring the mapper capacity $s_H$ is sufficient for robust cross-modal reconstruction without inducing overfitting.

\subsection{Prompt Capacity}
\label{sec:ctxseed}

Our investigation into prompt capacity, proxied by its Token Length (TL), reveals a critical trade-off between performance stability and representational expressiveness (Fig.~\ref{fig:pipeline}). A constrained capacity (TL=2) demonstrates remarkable training stability, with different random seeds converging to a tight performance cluster. In contrast, a higher capacity (TL=4) introduces significant performance variance, showing high sensitivity to initialization.

This dichotomy is well-explained by our \textit{point-to-cloud} thesis. A compact prompt (TL=2) enforces a more constrained search space, leading to a stable optimization landscape that facilitates reliable convergence. Conversely, a larger prompt (TL=4) creates a more expressive but complex optimization landscape, which, while offering a potentially higher performance ceiling, increases the risk of converging to suboptimal local minima.

\vspace{3mm}

%% file: sec/5_conclusion.tex
\section{Limitations and Future Work}
\label{sec:limitations}

The core principle of P2C appears to introduce a trade-off between generalization and specialization, as our denoising process acts as a strong regularizer. By smoothing the semantic landscape, it discourages overfitting to source-specific cues. While this enhances robustness against domain shifts, it may temper peak performance on the source domain or on closely related targets where more specialized representations can excel. This observation, coupled with the current reliance on a fixed GMM-based noise model, defines a clear path forward. Promising future directions include the development of adaptive noise mechanisms to better balance generalization and specialization, as well as the exploration of more efficient denoising strategies to enhance the framework's practicality and scalability.

\section{Conclusion}
\label{sec:conclusion}

In this work, we introduced \textit{Points-to-Clouds} (P2C), a framework that challenges the conventional static-point learning paradigm in multi-modal models. Inspired by diffusion models, P2C reframes prompt learning as a dynamic denoising task to learn a robust semantic cloud. Experiments demonstrate that P2C achieves 79.7\% harmonic mean on the challenging base-to-novel generalization benchmark, while also showing competitive robustness on domain generalization and cross-dataset transfer tasks. While promising, this paradigm introduces challenges such as hyperparameter sensitivity and increased training complexity, paving the way for future research.

%% file: sec/X_suppl.tex
\section*{Appendix.A.~Training Details}

\begin{figure*}[tbp]
    \centering
    \includegraphics[width=0.95\linewidth]{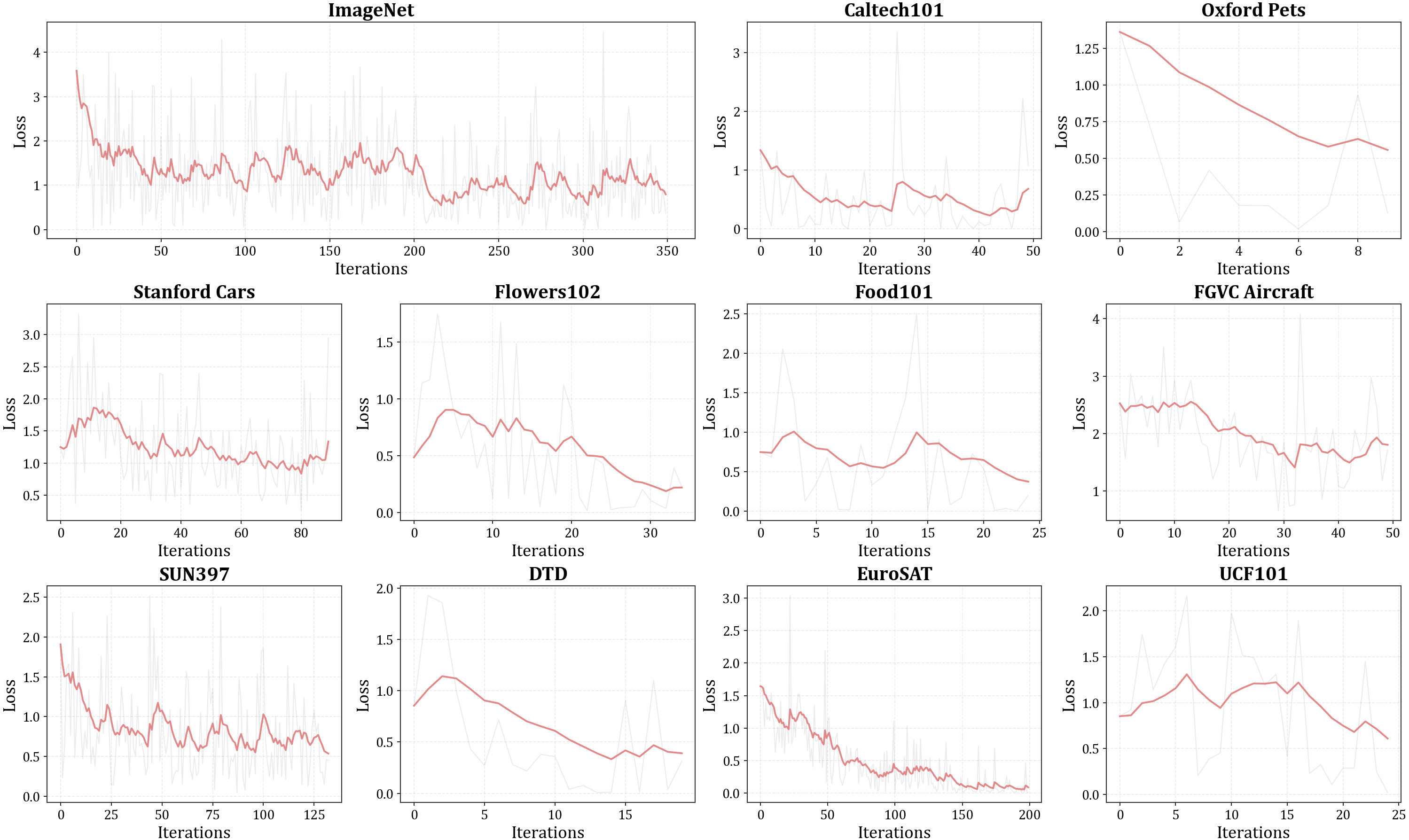}
    \caption{Convergence curves of our P2C trained on different datasets in the base-to-novel generalization task.}
    \label{fig:p2c_loss}
\end{figure*}

\begin{figure*}[tbp]
    \centering
    \includegraphics[width=0.95\linewidth]{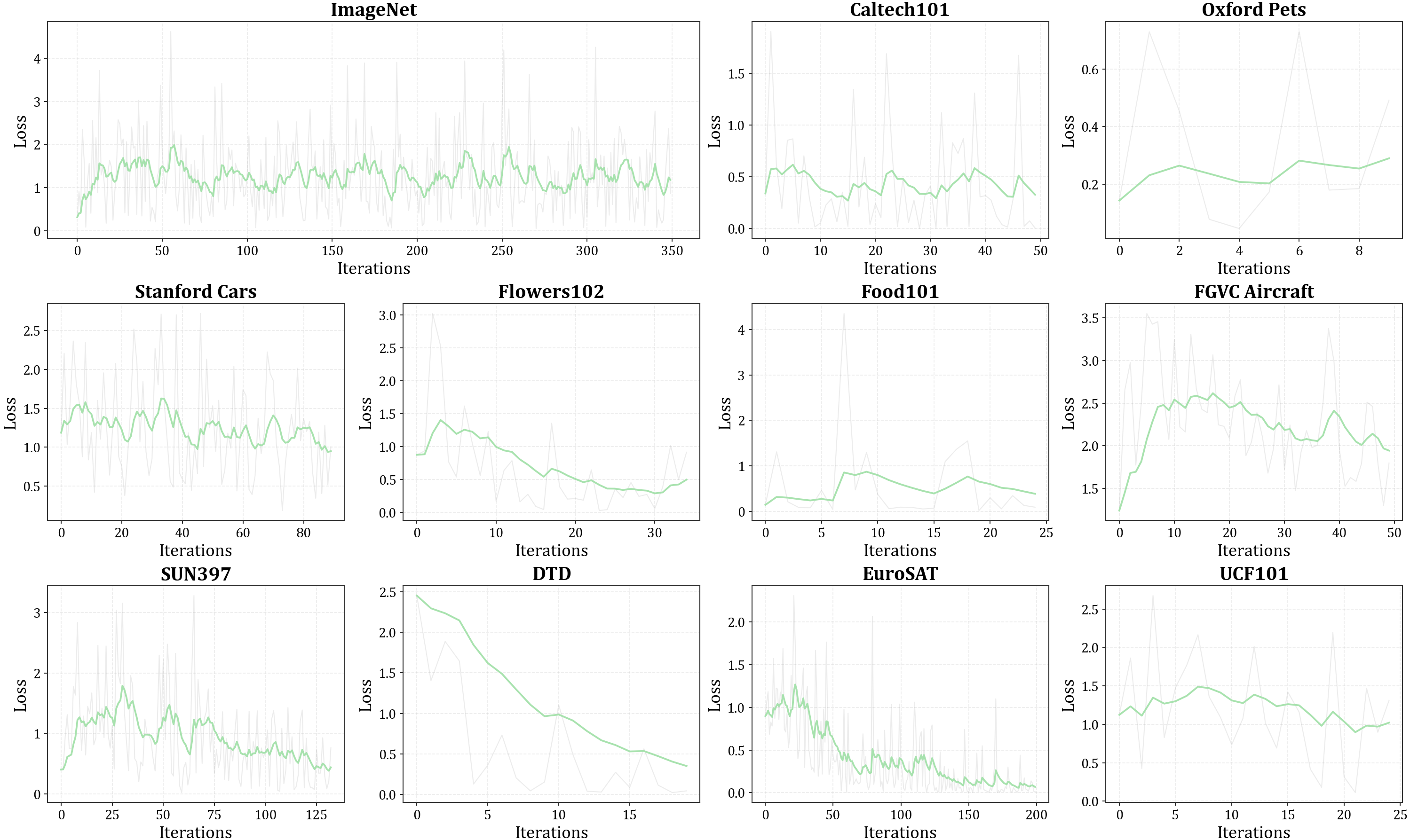}
    \caption{Convergence curves of the baseline method trained on different datasets in the base-to-novel generalization task.}
    \label{fig:baseline_loss}
\end{figure*}

In this section, we provide a comprehensive overview of the training dynamics and hyperparameter configurations used in our experiments. 

\textbf{Hyperparameter Configuration.} 
Following standard protocols, we employ the ViT-B/16 backbone for all experiments. The models are optimized using Stochastic Gradient Descent (SGD) with a batch size of 4. For the base-to-novel generalization benchmark, the learning rate is set to $3.5\times10^{-3}$, while for Cross-Dataset Transfer and Domain Generalization tasks, it is adjusted to $2.6\times10^{-3}$. The maximum noise standard deviation $\sigma_{max}$ for the Dynamic Prompt Denoising (DPD) is set to 0.015, and the weight $\lambda$ for the auxiliary denoising loss is set to 0.1, as given in our ablation studies.

\textbf{Convergence Analysis.} 
We visualize the training loss curves of our Points-to-Clouds (P2C) framework in Fig.~\ref{fig:p2c_loss} and compare them with the baseline method in Fig.~\ref{fig:baseline_loss}. Despite incorporating the DPD mechanism, which actively perturbs the input prompts, P2C still exhibits a highly stable convergence pattern across all 11 datasets.

As observed in Fig.~\ref{fig:p2c_loss}, the loss decreases consistently and stabilizes within the allocated iterations, similar to the baseline behavior shown in Fig.~\ref{fig:baseline_loss}. This demonstrates that our dual denoising mechanism acts as an effective regularizer without destabilizing the optimization landscape. The annealing schedule plays a crucial role here, ensuring that the noise magnitude is reduced as training progresses, allowing the model to fine-tune the semantic cloud representation smoothly.

\section*{Appendix.B.~Training Overhead}

\begin{figure*}[tbp]
    \centering
    \includegraphics[width=1\linewidth]{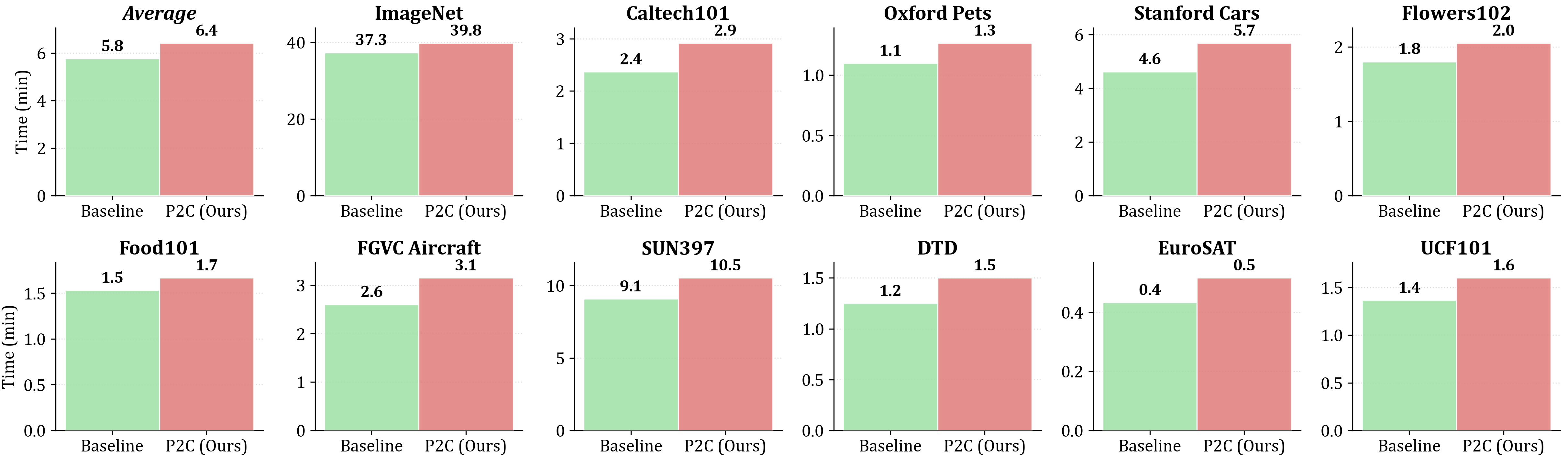}
    \caption{Comparison of training time between baseline methods and P2C on different datasets in the base-to-novel generalization task. P2C incurs only a marginal computational overhead during training.}
    \label{fig:fps}
\end{figure*}

We conduct a computational efficiency analysis to evaluate the cost of introducing the P2C framework. Fig.~\ref{fig:fps} presents a comparison of the training time per epoch between the baseline method and P2C across all evaluated datasets.

P2C incurs a slight increase in training duration compared to the baseline. This is attributed to the additional computational steps required for: (1) generating and injecting annealed noise into the prompts (DPD), and (2) computing the auxiliary V-L Mapper denoising loss for the V-L Mapper. However, as illustrated in Fig.~\ref{fig:fps}, this overhead is marginal (averaging approximately 6-25\% increase depending on the dataset size) and does not prohibitively affect the training feasibility.

\section*{Appendix.C.~Imbalance Assessment}

\begin{figure*}[tbp]
    \centering
    \includegraphics[width=1\linewidth]{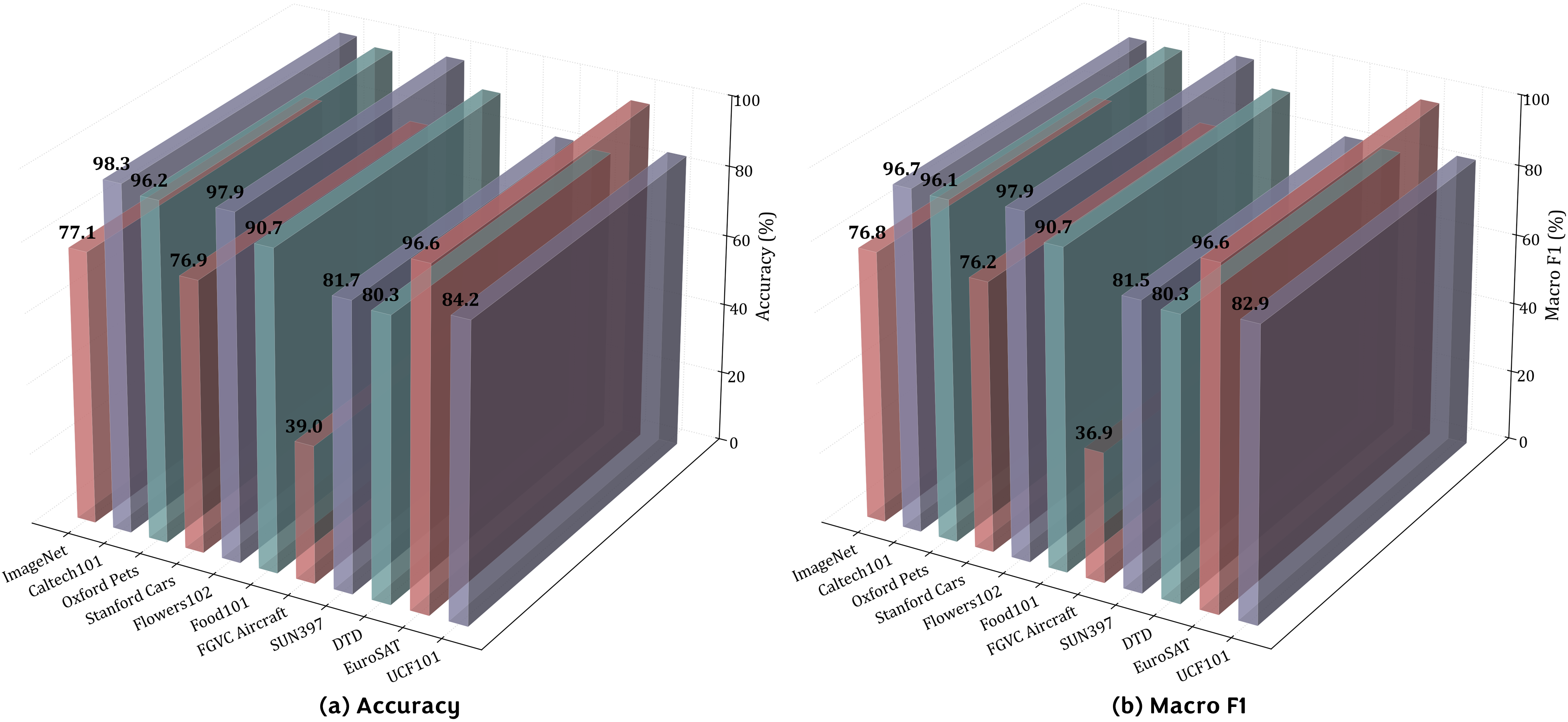}
    \caption{Imbalance Assessment on Base Classes. (a) P2C Accuracy performance confirms that our method preserves discriminative power on seen classes. (b) Consistently high Macro F1 scores indicate robustness against class imbalance within the base set, preventing decision boundary collapse.}
    \label{fig:imbalance_assessment}
\end{figure*}

A common pitfall in prompt learning, particularly in few-shot scenarios, is the tendency to overfit to the base classes, often leading to a degradation in base class performance when attempting to generalize to novel classes. Furthermore, real-world datasets often exhibit class imbalance, making it critical to maintain high precision and recall across all categories.

To assess P2C's robustness in this regard, we report both the Accuracy and Macro F1 scores specifically on the base classes in Fig.~\ref{fig:imbalance_assessment}.

\textbf{Preservation of Base Performance.}
As shown in Fig.~\ref{fig:imbalance_assessment}(a), P2C achieves high accuracy on base classes across diverse datasets, ranging from generic object recognition (ImageNet) to fine-grained tasks (FGVC Aircraft). This indicates that learning a semantic cloud does not dilute the discriminative power required for known categories.

\textbf{Robustness to Class Imbalance.}
Fig.~\ref{fig:imbalance_assessment}(b) illustrates the Macro F1 scores, which is a more robust metric for imbalanced data. The consistently high F1 scores (e.g., $>96\%$ on Caltech101, Oxford Pets, and EuroSAT) demonstrate that P2C handles class imbalance effectively. By optimizing for a distribution rather than a point, P2C prevents the decision boundaries from collapsing onto the majority samples of the base classes, thereby preserving the integrity of the semantic space for both base and novel categories.

%% file: main.bib
@String(AAAI = {AAAI})

@inproceedings{yao2024tcp,
  title={Tcp: Textual-based class-aware prompt tuning for visual-language model},
  author={Yao, Hantao and Zhang, Rui and Xu, Changsheng},
  booktitle={Proceedings of the IEEE/CVF Conference on Computer Vision and Pattern Recognition},
  pages={23438--23448},
  year={2024}
}

@inproceedings{khattak2023self,
  title={Self-regulating prompts: Foundational model adaptation without forgetting},
  author={Khattak, Muhammad Uzair and Wasim, Syed Talal and Naseer, Muzammal and Khan, Salman and Yang, Ming-Hsuan and Khan, Fahad Shahbaz},
  booktitle={Proceedings of the IEEE/CVF international conference on computer vision},
  pages={15190--15200},
  year={2023}
}

@inproceedings{li2024promptkd,
  title={Promptkd: Unsupervised prompt distillation for vision-language models},
  author={Li, Zheng and Li, Xiang and Fu, Xinyi and Zhang, Xin and Wang, Weiqiang and Chen, Shuo and Yang, Jian},
  booktitle={Proceedings of the IEEE/CVF Conference on Computer Vision and Pattern Recognition},
  pages={26617--26626},
  year={2024}
}

@inproceedings{zhu2023prompt,
  title={Prompt-aligned gradient for prompt tuning},
  author={Zhu, Beier and Niu, Yulei and Han, Yucheng and Wu, Yue and Zhang, Hanwang},
  booktitle={Proceedings of the IEEE/CVF international conference on computer vision},
  pages={15659--15669},
  year={2023}
}

@inproceedings{pan2025nlprompt,
  title={NLPrompt: Noise-Label Prompt Learning for Vision-Language Models},
  author={Pan, Bikang and Li, Qun and Tang, Xiaoying and Huang, Wei and Fang, Zhen and Liu, Feng and Wang, Jingya and Yu, Jingyi and Shi, Ye},
  booktitle={Proceedings of the Computer Vision and Pattern Recognition Conference},
  pages={19963--19973},
  year={2025}
}

@inproceedings{yang2024mma,
  title={Mma: Multi-modal adapter for vision-language models},
  author={Yang, Lingxiao and Zhang, Ru-Yuan and Wang, Yanchen and Xie, Xiaohua},
  booktitle={Proceedings of the IEEE/CVF Conference on Computer Vision and Pattern Recognition},
  pages={23826--23837},
  year={2024}
}

@inproceedings{khattak2023maple,
  title={Maple: Multi-modal prompt learning},
  author={Khattak, Muhammad Uzair and Rasheed, Hanoona and Maaz, Muhammad and Khan, Salman and Khan, Fahad Shahbaz},
  booktitle={Proceedings of the IEEE/CVF conference on computer vision and pattern recognition},
  pages={19113--19122},
  year={2023}
}

@inproceedings{yao2023visual,
  title={Visual-language prompt tuning with knowledge-guided context optimization},
  author={Yao, Hantao and Zhang, Rui and Xu, Changsheng},
  booktitle={Proceedings of the IEEE/CVF conference on computer vision and pattern recognition},
  pages={6757--6767},
  year={2023}
}

@inproceedings{zheng2025hierarchical,
  title={Hierarchical cross-modal prompt learning for vision-language models},
  author={Zheng, Hao and Yang, Shunzhi and He, Zhuoxin and Yang, Jinfeng and Huang, Zhenhua},
  booktitle={Proceedings of the IEEE/CVF International Conference on Computer Vision},
  pages={1891--1901},
  year={2025}
}

@inproceedings{xu2025fate,
  title={FATE: Feature-Adapted Parameter Tuning for Vision-Language Models},
  author={Xu, Zhengqin and Peng, Zelin and Yang, Xiaokang and Shen, Wei},
  booktitle={Proceedings of the AAAI Conference on Artificial Intelligence},
  volume={39},
  number={9},
  pages={9014--9022},
  year={2025}
}

@inproceedings{li2025dpc,
  title={Dpc: Dual-prompt collaboration for tuning vision-language models},
  author={Li, Haoyang and Wang, Liang and Wang, Chao and Jiang, Jing and Peng, Yan and Long, Guodong},
  booktitle={Proceedings of the Computer Vision and Pattern Recognition Conference},
  pages={25623--25632},
  year={2025}
}

@inproceedings{li2025divergence,
  title={Divergence-enhanced knowledge-guided context optimization for visual-language prompt tuning},
  author={Li, Yilun and Cheng, Miaomiao and Han, Xu and Song, Wei},
  booktitle={The Thirteenth International Conference on Learning Representations},
  year={2025}
}

@article{zhang2025decouple,
  title={Decouple before align: Visual disentanglement enhances prompt tuning},
  author={Zhang, Fei and Zhou, Tianfei and Yao, Jiangchao and Zhang, Ya and Tsang, Ivor W and Wang, Yanfeng},
  journal={IEEE Transactions on Pattern Analysis and Machine Intelligence},
  year={2025},
  publisher={IEEE}
}

@article{zhou2022learning,
  title={Learning to prompt for vision-language models},
  author={Zhou, Kaiyang and Yang, Jingkang and Loy, Chen Change and Liu, Ziwei},
  journal={International Journal of Computer Vision},
  volume={130},
  number={9},
  pages={2337--2348},
  year={2022},
  publisher={Springer}
}

@inproceedings{zhou2022conditional,
  title={Conditional prompt learning for vision-language models},
  author={Zhou, Kaiyang and Yang, Jingkang and Loy, Chen Change and Liu, Ziwei},
  booktitle={Proceedings of the IEEE/CVF conference on computer vision and pattern recognition},
  pages={16816--16825},
  year={2022}
}

@inproceedings{radford2021learning,
  title={Learning transferable visual models from natural language supervision},
  author={Radford, Alec and Kim, Jong Wook and Hallacy, Chris and Ramesh, Aditya and Goh, Gabriel and Agarwal, Sandhini and Sastry, Girish and Askell, Amanda and Mishkin, Pamela and Clark, Jack and others},
  booktitle={International conference on machine learning},
  pages={8748--8763},
  year={2021},
  organization={PmLR}
}

@inproceedings{wu2024cascade,
  title={Cascade prompt learning for vision-language model adaptation},
  author={Wu, Ge and Zhang, Xin and Li, Zheng and Chen, Zhaowei and Liang, Jiajun and Yang, Jian and Li, Xiang},
  booktitle={European Conference on Computer Vision},
  pages={304--321},
  year={2024},
  organization={Springer}
}

@article{yao2025bi,
  title={Bi-modality Individual-aware Prompt tuning for Visual-Language Model},
  author={Yao, Hantao and Zhang, Rui and Lyu, Huaihai and Zhang, Yongdong and Xu, Changsheng},
  journal={IEEE Transactions on Pattern Analysis and Machine Intelligence},
  year={2025},
  publisher={IEEE}
}

@article{li2024atprompt,
  title={ATPrompt: Textual Prompt Learning with Embedded Attributes},
  author={Li, Zheng and Song, Yibing and Zhao, Penghai and Cheng, Ming-Ming and Li, Xiang and Yang, Jian},
  journal={arXiv preprint arXiv:2412.09442},
  year={2024}
}

@article{ho2020denoising,
  title={Denoising diffusion probabilistic models},
  author={Ho, Jonathan and Jain, Ajay and Abbeel, Pieter},
  journal={Advances in neural information processing systems},
  volume={33},
  pages={6840--6851},
  year={2020}
}

@inproceedings{deng2009imagenet,
  title={Imagenet: A large-scale hierarchical image database},
  author={Deng, Jia and Dong, Wei and Socher, Richard and Li, Li-Jia and Li, Kai and Fei-Fei, Li},
  booktitle={2009 IEEE conference on computer vision and pattern recognition},
  pages={248--255},
  year={2009},
  organization={Ieee}
}

@inproceedings{fei2004learning,
  title={Learning generative visual models from few training examples: An incremental bayesian approach tested on 101 object categories},
  author={Fei-Fei, Li and Fergus, Rob and Perona, Pietro},
  booktitle={2004 conference on computer vision and pattern recognition workshop},
  pages={178--178},
  year={2004},
  organization={IEEE}
}

@inproceedings{parkhi2012cats,
  title={Cats and dogs},
  author={Parkhi, Omkar M and Vedaldi, Andrea and Zisserman, Andrew and Jawahar, CV},
  booktitle={2012 IEEE conference on computer vision and pattern recognition},
  pages={3498--3505},
  year={2012},
  organization={IEEE}
}

@inproceedings{krause20133d,
  title={3d object representations for fine-grained categorization},
  author={Krause, Jonathan and Stark, Michael and Deng, Jia and Fei-Fei, Li},
  booktitle={Proceedings of the IEEE international conference on computer vision workshops},
  pages={554--561},
  year={2013}
}

@inproceedings{nilsback2008automated,
  title={Automated flower classification over a large number of classes},
  author={Nilsback, Maria-Elena and Zisserman, Andrew},
  booktitle={2008 Sixth Indian conference on computer vision, graphics \& image processing},
  pages={722--729},
  year={2008},
  organization={IEEE}
}

@inproceedings{bossard2014food,
  title={Food-101--mining discriminative components with random forests},
  author={Bossard, Lukas and Guillaumin, Matthieu and Van Gool, Luc},
  booktitle={European conference on computer vision},
  pages={446--461},
  year={2014},
  organization={Springer}
}

@article{maji2013fine,
  title={Fine-grained visual classification of aircraft},
  author={Maji, Subhransu and Rahtu, Esa and Kannala, Juho and Blaschko, Matthew and Vedaldi, Andrea},
  journal={arXiv preprint arXiv:1306.5151},
  year={2013}
}

@inproceedings{xiao2010sun,
  title={Sun database: Large-scale scene recognition from abbey to zoo},
  author={Xiao, Jianxiong and Hays, James and Ehinger, Krista A and Oliva, Aude and Torralba, Antonio},
  booktitle={2010 IEEE computer society conference on computer vision and pattern recognition},
  pages={3485--3492},
  year={2010},
  organization={IEEE}
}

@article{soomro2012ucf101,
  title={Ucf101: A dataset of 101 human actions classes from videos in the wild},
  author={Soomro, Khurram and Zamir, Amir Roshan and Shah, Mubarak},
  journal={arXiv preprint arXiv:1212.0402},
  year={2012}
}

@inproceedings{cimpoi2014describing,
  title={Describing textures in the wild},
  author={Cimpoi, Mircea and Maji, Subhransu and Kokkinos, Iasonas and Mohamed, Sammy and Vedaldi, Andrea},
  booktitle={Proceedings of the IEEE conference on computer vision and pattern recognition},
  pages={3606--3613},
  year={2014}
}

@article{helber2019eurosat,
  title={Eurosat: A novel dataset and deep learning benchmark for land use and land cover classification},
  author={Helber, Patrick and Bischke, Benjamin and Dengel, Andreas and Borth, Damian},
  journal={IEEE Journal of Selected Topics in Applied Earth Observations and Remote Sensing},
  volume={12},
  number={7},
  pages={2217--2226},
  year={2019},
  publisher={IEEE}
}

@inproceedings{recht2019imagenet,
  title={Do imagenet classifiers generalize to imagenet?},
  author={Recht, Benjamin and Roelofs, Rebecca and Schmidt, Ludwig and Shankar, Vaishaal},
  booktitle={International conference on machine learning},
  pages={5389--5400},
  year={2019},
  organization={PMLR}
}

@article{wang2019learning,
  title={Learning robust global representations by penalizing local predictive power},
  author={Wang, Haohan and Ge, Songwei and Lipton, Zachary and Xing, Eric P},
  journal={Advances in neural information processing systems},
  volume={32},
  year={2019}
}

@inproceedings{hendrycks2021natural,
  title={Natural adversarial examples},
  author={Hendrycks, Dan and Zhao, Kevin and Basart, Steven and Steinhardt, Jacob and Song, Dawn},
  booktitle={Proceedings of the IEEE/CVF conference on computer vision and pattern recognition},
  pages={15262--15271},
  year={2021}
}

@inproceedings{hendrycks2021many,
  title={The many faces of robustness: A critical analysis of out-of-distribution generalization},
  author={Hendrycks, Dan and Basart, Steven and Mu, Norman and Kadavath, Saurav and Wang, Frank and Dorundo, Evan and Desai, Rahul and Zhu, Tyler and Parajuli, Samyak and Guo, Mike and others},
  booktitle={Proceedings of the IEEE/CVF international conference on computer vision},
  pages={8340--8349},
  year={2021}
}
